\documentclass{article}
\usepackage[english]{babel}
\usepackage[utf8]{inputenc}
\usepackage{johd}
\usepackage{amssymb}
\usepackage{amsmath}
\usepackage{pbox}
\usepackage{hyperref}

\title{Physics-informed PointNet: On how many irregular geometries can it solve an inverse problem simultaneously? Application to linear elasticity}

\author{Ali Kashefi $^{a}$$^{*}$, Leonidas J. Guibas $^{b}$, Tapan Mukerji $^{c}$  \\
        \small $^{a}$Department of Civil and Environmental Engineering, Stanford University, Stanford, CA, 94305 USA \\
        \small $^{b}$Department of Computer Science, Stanford University, Stanford, CA, 94305 USA \\
        \small $^{c}$Department of Energy Science and Engineering, Stanford University, Stanford, CA, 94305 USA \\
 \\\\
        \small $^{*}$Corresponding author: Ali Kashefi, \tt{ kashefi@stanford.edu}
}

\date{} %leave blank

\begin{document}

%\counterwithout*{figure}{section}

\maketitle

\begin{abstract} 
\noindent Regular physics-informed neural networks (PINNs) predict the solution of partial differential equations using sparse labeled data but only over a single domain. On the other hand, fully supervised learning models are first trained usually over a few thousand domains with known solutions (i.e., labeled data) and then predict the solution over a few hundred unseen domains. Physics-informed PointNet (PIPN) is primarily designed to fill this gap between PINNs (as weakly supervised learning models) and fully supervised learning models. In this article, we demonstrate for the first time that PIPN predicts the solution of desired partial differential equations over a few hundred domains simultaneously, while it only uses sparse labeled data. This framework benefits fast geometric designs in the industry when only sparse labeled data are available. Particularly, we show that PIPN predicts the solution of a plane stress problem over more than 500 domains with different geometries, simultaneously. Moreover, we pioneer implementing the concept of remarkable batch size (i.e., the number of geometries fed into PIPN at each sub-epoch) into PIPN. We systematically try batch sizes of 7, 14, 19, 38, 76, and 133. Additionally, we systematically investigate for the first time the effect of the PIPN size, symmetric function in the PIPN architecture, and static and dynamic weights for the component of the sparse labeled data in the PIPN loss function. 
\end{abstract}

\noindent\\

%\noindent\authorroles{For determining author roles, please use following taxonomy: \url{https://credit.niso.org/}. Please list the roles for each author.} 

\section*{Highlights}

\begin{enumerate}

  \item Similar to supervised deep learning, for the first time we test big data (i.e., number of geometry) and relatively large batch sizes (i.e., number of geometries per epoch) for the framework of physics-informed neural networks.
  
  \item We specifically use PIPN as an advanced version of physics-informed neural networks.
  
  \item Using PIPN, an inverse problem of linear elasticity is solved over 532 irregular geometries simultaneously.
  
  \item Batch sizes (i.e., number of geometries per epoch) of 7,14, 19, 28, 38, 76, and 133 are tested.

   \item The effect of the PIPN size and symmetric function in the PIPN architecture are explored for the first time.
  
\end{enumerate}

\section{Introduction and motivation}\label{Sect1}

Physics-informed Neural Networks (PINNs), introduced by Raissi, et al. \cite{raissi2019physics} in 2019, are recognized as a promising tool for solving inverse problems in a variety of scientific and industrial fields such as solid mechanics \cite{haghighat2021physics,rao2021physics,jiang2022physics,tandale2022physics,vadyala2021review,xu2023transfer,flaschel2022discovering,cao2022physics,wu2022effective,bai2022physics,he2023mflp,bolandi2022physics,niu2023modeling,rezaei2022mixed,jeong2023physics,qiu2023sensenet,fernandez2023physics}, incompressible and compressible flows \cite{jin2021nsfnets,kashefi2022physics,lou2021physics,jagtap2022physics,jagtap2020conservative,raissi2019physics,patel2022thermodynamically,qiu2022physics,ouyang2023reconstruction,buhendwa2021inferring}, chemistry \cite{weng2022multiscale,ji2021stiff}, heat transfer \cite{kashefi2022physics,wang2021reconstruction,cai2021physics}, flow in porous media \cite{kashefi2022prediction,almajid2022prediction,yu2022gradient}, etc. The main idea of PINNs for solving inverse problems can be briefly explained as follows. Given sparse observations of a field of interest as well as the partial differential equations (PDEs) governing the physics of the field, train ``a neural network'' such that its predictions (outputs) minimize the residuals of the PDEs as well as the distance between the predictions and sparse observations at sensor locations, in a certain norm (such as $L^2$ norm or other norms). Other partial information such as boundary or initial conditions may be included in this minimization problem. Our focus, here, is on the term ``a neural network''. We believe that the choice of the neural network significantly affects the ability and capacity of a PINN configuration. A common choice is a fully connected neural network (e.g., see Fig. 1 in Ref. \cite{rao2021physics}, Fig. 3 in Ref. \cite{haghighat2021physics}, Fig. 2 in Ref. \cite{xu2021explore}, Fig. 1 in Ref. \cite{lou2021physics}, Fig. 3 in Ref. \cite{yuan2022pinn}, Fig 1 in Ref. \cite{yang2021bayesian}, Fig. 1 in Ref. \cite{mao2020physics}, Fig. 4 in Ref. \cite{linka2022bayesian}, Fig. 1 in Ref. \cite{eivazi2022physics}, etc.). An immediate consequence of this choice is that a PINN with a fully connected neural network is only able to predict the solution of an inverse problem on a single geometry. Hence, for any computational domain with a new geometry, one needs to train a PINN from scratch. This scenario necessitates high computational expenses, specifically when the goal is the investigation of a wide range of geometric parameters for optimizing an industrial design. This issue has been first addressed by Gao, et al. \cite{gao2021phygeonet} in 2021 and later by Kashefi and Mukerji \cite{kashefi2022physics} in 2022.

\begin{figure*}
\label{Fig1}
\centering
\includegraphics[width=1.0\textwidth]{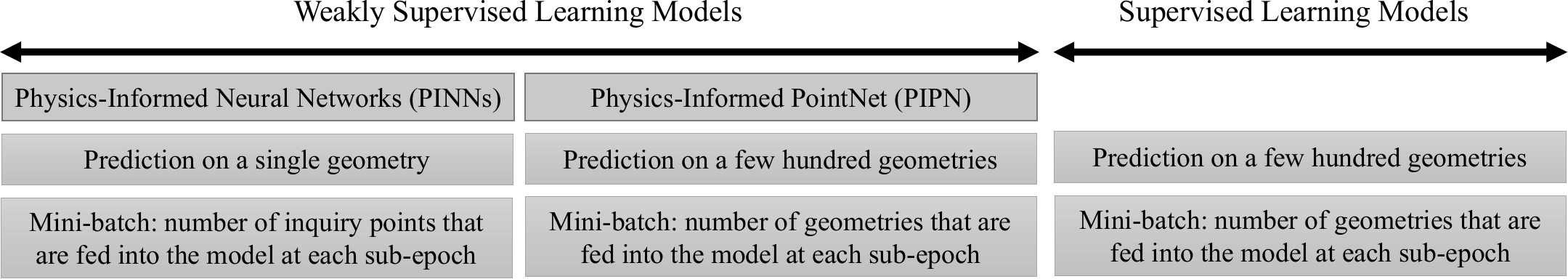}
\caption{A conceptual comparison between regular physics-informed neural networks (PINNs), physics-informed PointNet (PIPN), and fully supervised deep learning algorithms for computational mechanics in terms of the number of geometries and the batch size definition}
\end{figure*}

To resolve this issue, Gao, et al. \cite{gao2021phygeonet} first proposed the PhyGeoNet \cite{gao2021phygeonet} framework and later Kashefi and Mukerji \cite{kashefi2022physics} introduced the PIPN \cite{kashefi2022physics} framework. PIPN \cite{kashefi2022physics} also successfully overcame the shortcomings of PhyGeoNet \cite{gao2021phygeonet}. We later compare PIPN \cite{kashefi2022physics} versus PhyGeoNet \cite{gao2021phygeonet}, but for now, let us focus on the main theme of our article. What was one of the main motivations for introducing PIPN? To answer this question, let us discuss PINNs from another perspective. PINNs can be categorized as weakly supervised learning frameworks in the sense that the associated neural networks predict the solution on a computational domain using some sparse labeled data, which are available on that specific domain. On the other hand, neural networks used in supervised learning methods for computational mechanics are first trained on a large number of computational domains with different geometries, where the solutions are known, and then they predict the solution on new computational domains with new geometries. Supervised learning methods are commonly trained on a few hundred to a few thousand geometries (e.g., 2595 in Ref. \cite{kashefi2021point}, 1505 in Ref. \cite{thuerey2019deep}, 880 in Ref. \cite{sekar2019fast}, 1600 in Ref. \cite{chen2019aerodynamic}, etc.). One of the goals of introducing PIPN was to fill the gap between the weakly-supervised learning and supervised learning models in terms of the number of geometries for which we can predict the solution simultaneously (see Fig. \ref{Fig1}). Now, the question is given the current capacity of available commercial graphics processing units (GPUs) to the public, how many computational domains (with different geometries) is PIPN able to predict the solution of an inverse problem on, simultaneously?

The answer to this question depends on different parameters. But definitely one of them is the level of complexity of a problem, which has a direct relevancy to its governing PDEs. When Kashefi and Mukerji \cite{kashefi2022physics} introduced PIPN, they \cite{kashefi2022physics} showed its applications for incompressible flows and natural convection, where the governing PDEs are nonlinear (see Eqs. 1--3 of Ref. \cite{kashefi2022physics}). Due to the nonlinearity of the problem, they had to select high resolutions (number of inquiry points) per computational domain. Specifically for the natural convection problem (see Sect. 4.2 of Ref. \cite{kashefi2022physics}), they run PIPN on 108 geometries, while each geometry had a resolution of 5000 inquiry points. They \cite{kashefi2022physics} further showed the effect of resolution on the accuracy of the PIPN prediction (see Fig. 18 of Ref. \cite{kashefi2022physics}). Because of the current limitations on GPU memories, they were not able to explore a ``big'' data set comparable to those used in supervised learning. One way to reduce the complexity level is to select a set of linear PDEs to examine the PIPN capacity. A reasonable choice is the equations of two-dimensional linear elasticity problems, which have a wide range of applications in the industry. Hence, to examine the capacity of PIPN in terms of simultaneous handling of the maximum possible number of geometries, we focus on the two-dimensional linear elasticity problems. Additionally, it is worth noting that this is the first time that PIPN is used for solving an inverse problem of linear elasticity.

An important hyperparameter that needs to be tuned for training neural networks in supervised learning of computational mechanics is ``batch size''. The concept of ``batch size'' is defined under the category of mini-batch gradient descent \cite{goodfellow2016deep}, where the training data set is split into smaller data sets as mini-batches. The size of each mini-batch is then called ``batch size''. The batch size plays a critical role in the training convergence as well as the ability of a trained neural network to generalize predictions \cite{kandel2020effect,keskar2016large,bengio2012practical,masters2018revisiting}. In supervised deep learning of quantities of interest as a function of geometric features of domains, the term batch size refers to the number of domains, with different geometries, that are fed to a neural network at each epoch. In regular PINNs, however, the term batch size refers to the number of inquiry points belonging to a sub-domain of a single domain that is fed to a neural network at each epoch. In PIPN, Kashefi and Mukerji \cite{kashefi2022physics} used the term batch size to address the number of domains at each epoch, similar to the scenario for supervised learning. Nevertheless, the maximum batch size was 13 as reported in their research article \cite{kashefi2022physics}, again due to the GPU memory limitations and nonlinearity of their considered PDEs. In this article, we investigate for the first time a wide range of batch sizes and their influences on the training convergence and generalizability of PIPN with its applications to a linear elasticity problem.

Furthermore, we study for the first time another important hyperparameter, which is the network size of PIPN. In the framework of regular PINNs, where fully connected layers are used, the network size is tuned by changing the number of layers and a number of neurons in each layer. Contrarily to regular PINNs, PIPN has a more complicated architecture, and its network size can practically vary in different ways. In this work, we define a global scalar variable to control the network size of PIPN. Afterward, we study the performance of the PIPN framework as a function of this global variable.

Moreover, we investigate for the first time the effect of static and dynamic weights of the different components of the PIPN loss function on the PIPN performance in terms of both convergence rate and prediction accuracy. In addition, we explore for the first time the influence of symmetric functions, embedded in the PIPN architecture, on the accuracy of the PIPN outputs.

For the sake of completeness and also convincing the potential audiences why we go with PIPN \cite{kashefi2022physics} rather than PhyGeoNet \cite{gao2021phygeonet} in this article, we compare PIPN \cite{kashefi2022physics} with PhyGeoNet \cite{gao2021phygeonet}. In PhyGeoNet \cite{gao2021phygeonet}, a convolutional neural network (CNN) is used as the ``neural network'' in a PINN, instead of a fully connected neural network. CNNs are able to extract geometric features of input computational domains and represent those features in a hidden space. Hence, all the parameters of CNNs (e.g., weights and bias) become a function of the geometric features. As a result, PhyGeoNet \cite{gao2021phygeonet} is able to solve an inverse problem on multiple sets of computational domains. Notwithstanding these successes, PhyGeoNet \cite{gao2021phygeonet} and its later version \cite{ren2022phycrnet} come with several shortcomings. These shortcomings have been addressed in detail in Sect. 1 of Ref. \cite{kashefi2022physics}. Here, we summarize them. First, PhyGeoNet \cite{gao2021phygeonet} uses a finite difference discretization method for computing the loss function of desired PDEs rather than using the automatic differentiation technology \cite{tensorflow2015-whitepaper}. In this way, PhyGeoNet \cite{gao2021phygeonet} is limited to the accuracy order of the chosen finite difference scheme and faces challenges for treating near boundaries of computational domains in the case of using high-order finite difference methods. Second, PhyGeoNet \cite{gao2021phygeonet} is not able to handle irregular ``non-parameterizable'' geometries. Third, even on irregular parameterizable geometries, PhyGeoNet \cite{gao2021phygeonet} is strictly limited to handling domains with up to only five $C_0$ continuous boundaries.

To obviate the limitations of PhyGeoNet \cite{gao2021phygeonet}, Kashefi and Mukerji \cite{kashefi2022physics} introduced physics-informed PointNet (PIPN). In PIPN, PointNet \cite{qi2017pointnet} carries out the role of the ``neural network'' in PINNs. Similar to CNNs, PointNet \cite{qi2017pointnet} is able to also encode the geometric features of input computational domains; however, using a different mathematical methodology. Introduced in 2017, PointNet \cite{qi2017pointnet} emerged as a prominent deep learning model for addressing 3D computer vision problems such as shape classification and segmentation. PointNet \cite{qi2017pointnet} set the stage for subsequent research, such as object detection in both outdoor \cite{qi2018frustum} and indoor \cite{qi2019deep} environments, and the prediction of scene flow from sequential point cloud data \cite{liu2019flownet3d}. The classification branch of PointNet \cite{qi2017pointnet} was also used for identifying structure in molecular simulations \cite{defever2019generalized} and atomistic simulations \cite{shen2021development} as well as predicting porous medium permeability \cite{kashefi2021pointPorous}. Although advanced versions of PointNet have surfaced, with adaptations like hierarchical PointNet \cite{qi2017pointnet++} and innovative approaches utilizing graph \cite{wang2019dynamic} or continuous convolutions \cite{thomas2019kpconv} on point clouds, the original PointNet \cite{qi2017pointnet} continues to stand out for its conceptual and computational simplicity. Details of the PointNet architecture can be found in Refs. \cite{qi2017pointnet,kashefi2021point,kashefi2022physics,kashefi2021pointPorous}. We also briefly review the PointNet structure in Sect. \ref{Sect22} of this research paper. The advantages of PIPN compared to PhyGeoNet have been explained in detail by Kashefi and Mukerji \cite{kashefi2022physics}. We list them here for a quick review. First and most important, PIPN \cite{kashefi2022physics} can handle non-parametrizable irregular geometries without any restriction. Second, PIPN \cite{kashefi2022physics} uses automatic differentiation technology \cite{tensorflow2015-whitepaper} and thus the spatial derivative of output components with respect to the corresponding input component can be conveniently programmed and computed over entire irregular geometries in a data set (with no need for a finite difference/element discretization). Third, in the PIPN framework, interior and boundary points of irregular geometries are explicitly identified, and it allows twofold benefits: first, a smooth representation of boundaries (e.g., an airfoil surface), and second, explicit identification of their corresponding components in the loss function (with no need for implicit labeling of boundary points versus interior points and an artificial mixing up near boundaries, especially with sharp corners).

The rest of this research paper is structured as follows. We mathematically formulate the linear elasticity problem in Sect. \ref{Sect21}. An illustration of PIPN from computer science and applied mathematics perspectives are given in Sect. \ref{Sect22}. Details about generating data for validation of the PIPN framework are described in Sect. \ref{Sect23}. We elaborate on the computational setup for training PIPN in Sect. \ref{Sect24}. A general analysis of the PIPN performance as well as the effect of batch size and neural network size is discussed in Sect. \ref{Sect3}. Concluding remarks are listed in Sect. \ref{Sect4}.

\section{Problem statement and methodologies}\label{Sect2}

\subsection{Problem formulations}\label{Sect21}

In this article, we focus on elastic materials with the constitutive model of Hooke's law, relating the Cauchy stress tensor to the infinitesimal strain tensor and we restrict our studies to isotropic materials under plane stress assumptions. Specifically, we consider a two-dimensional domain $V$ having a cavity space characterized by various shapes. The domain $V$ is placed under a thermal conductivity loading. The static linear elasticity equations governing the displacement fields in the domain $V$ are given by

\begin{equation}
\label{Eq1}
-\frac{\partial}{\partial x}\Bigl( \frac{E}{1-\nu^2}\frac{\partial u}{\partial x}+\frac{E\nu}{1-\nu^2}\frac{\partial v}{\partial y}\Bigl) -\frac{\partial}{\partial y}\Bigl(\frac{E}{2(1+\nu)}\bigl(\frac{\partial u}{\partial y} + \frac{\partial v}{\partial x}\bigl)\Bigl) = -\frac{E \alpha}{1-\nu^2}\frac{\partial T}{\partial x}, 
\end{equation}

\begin{equation}
\label{Eq2}
-\frac{\partial}{\partial y}\Bigl( \frac{E \nu}{1-\nu^2}\frac{\partial u}{\partial x}+\frac{E}{1-\nu^2}\frac{\partial v}{\partial y}\Bigl) -\frac{\partial}{\partial x}\Bigl(\frac{E}{2(1+\nu)}\bigl(\frac{\partial u}{\partial y} + \frac{\partial v}{\partial x}\bigl)\Bigl) = - \frac{E \alpha}{1-\nu^2}\frac{\partial T}{\partial y}, 
\end{equation}
where $T$ indicates the temperature variable. Displacements in the $x$ and $y$ directions are shown by $u$ and $v$, respectively. $\alpha$ denotes the thermal expansion coefficient. $E$ is the elastic Young’s modulus and $\nu$ is the Poisson ratio. We assume that the material of the domain $V$ remains in the elastic regime under thermal loading.

Mathematically, our goal is to solve an inverse problem using PIPN on a set of irregular domains $\Phi=\{V_i\}_{i=1}^m$, formulated as follows: given the temperature field of the domains $\Phi=\{V_i\}_{i=1}^m$ and a set of sparse observations of the displacement fields of the domains of the set $\Phi=\{V_i\}_{i=1}^m$, find the full displacement solutions for all the domains of the set $\Phi=\{V_i\}_{i=1}^m$. We explicitly illustrate geometric variations of the set $\Phi=\{V_i\}_{i=1}^m$ in Sect. \ref{Sect23}.

\subsection{Physics-informed PointNet (PIPN)}\label{Sect22}

As we discussed in Sect. \ref{Sect1}, the idea of PIPN was first proposed by Kashefi and Mukerji \cite{kashefi2022physics} in 2022. Here, we briefly review the PIPN methodology and adjust it for the governing equations of our interest (Eqs. \ref{Eq1}--\ref{Eq2}). PIPN is built based on the combination of two critical pieces: PointNet \cite{qi2017pointnet} and a physics-based loss function. In simple words, the PIPN mechanism can be explained in two steps. In the first step, PointNet \cite{qi2017pointnet} makes the predicted outputs as a function of geometric features of each $V_i \in \Phi$. As the second step, the PIPN loss function is computed by taking these outputs and manipulating them using automatic differentiation \cite{tensorflow2015-whitepaper} to build up the equations describing the physics of the problem. Hence, the PIPN loss function is not only aware of the physics but also aware of the geometries of each $V_i \in \Phi$. In this way, by training PIPN, the predicted solutions for the displacement fields on a domain $V_i$ become a function of the geometric characteristic of $V_i$. In this sense, PIPN is able to predict the solutions of the governing PDEs (Eqs. \ref{Eq1}--\ref{Eq2}) on multiple computational domains with various geometries, simultaneously.

\subsubsection{Architecture}
\label{Sect221}

\begin{figure*}
\centering
\includegraphics[width=1.0\textwidth]{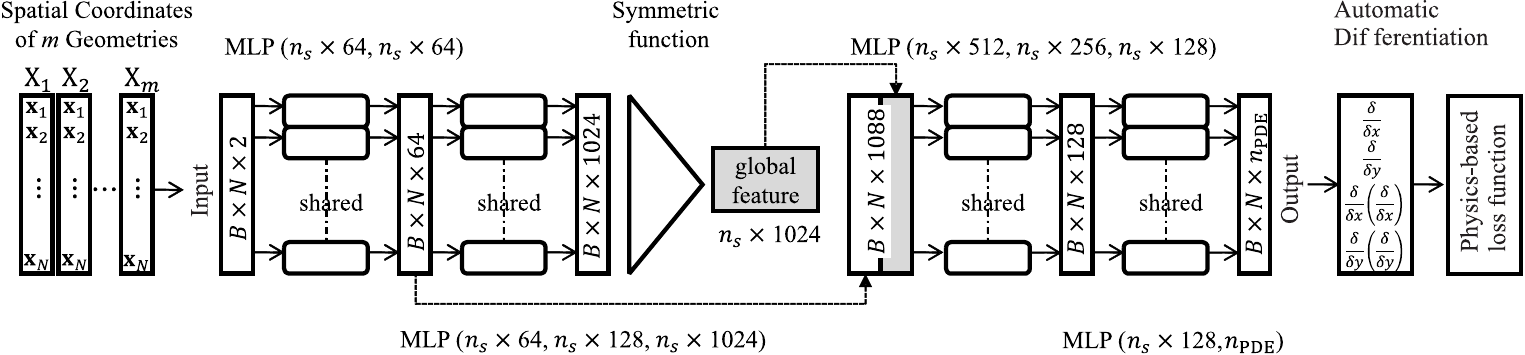}
\caption{Architecture of physics-informed PointNet (PIPN); $B$ indicates the batch size and $n_s$ is a global scaling parameter for controlling the network size. The spatial derivatives are computed using automatic differentiation to build up the residuals of the governing PDEs as the PIPN loss function.}
\label{Fig2}
\end{figure*}

The PIPN architecture is exhibited in Fig. \ref{Fig2}. In PIPN, each $V_i$ is represented by a point-cloud $\mathcal{X}_i$ with $N$ points. $\mathcal{X}_i$ is defined as $\mathcal{X}_i=\{\textbf{x}_j\in \mathbb{R}^{d}\}_{j=1}^N$, where $d$ is the spatial dimension and we set $d=2$ in this research study. Thus, each $\textbf{x}_j$ has two components ${x}_j$ and ${y}_j$ as the spatial coordinates. PIPN maps $\mathcal{X}_i$ to $\mathcal{Y}_i$ via a function $f$, where $\mathcal{Y}_i$ is the prediction of the PDE (Eqs. \ref{Eq1}--\ref{Eq2}) solutions. $\mathcal{Y}_i$ is defined as $\mathcal{Y}_i=\{\textbf{y}_j\in \mathbb{R}^{n_{\text{PDE}}}\}_{j=1}^N$, where $n_{\text{PDE}}$ indicates the number of fields in the solution. Here, we set $n_{\text{PDE}}=2$ as the displacement fields ($u$ and $v$) are the unknowns. Hence, each $\textbf{y}_j$ has two components, $u_j$ and $v_j$ as the network outputs. Mathematically, it can be written as

\begin{equation}
\label{Eq3}
(u_j,v_j) = f((x_j,y_j),g(\mathcal{X}_i)); \forall (x_j,y_j)\in \mathcal{X}_i \text{ and } \forall (u_j,v_j) \in \mathcal{Y}_i  \text{ with } 1 \leq i \leq m \text{ and } 1 \leq j \leq N, 
\end{equation}
where $f$ is the mapping function approximated by the PointNet \cite{qi2017pointnet} neural network and $g$ is a symmetric function representing the geometric feature of a point cloud $\mathcal{X}_i$ and can be approximated as

\begin{equation}
\label{Eq4}
    g(\mathcal{X}_i)=s(h(x_1,y_1 ),\dots,h(x_N,y_N)); \forall (x_j,y_j)\in \mathcal{X}_i \text{ with } 1 \leq i \leq m \text{ and } 1 \leq j \leq N,   
\end{equation}
where $s$ is a symmetric function and $h$ is a function representing the two shared Multilayer Perceptrons (MLPs) \cite{lin2013network,qi2017pointnet,qi2017pointnet++} in the first branch of PointNet \cite{qi2017pointnet} (see Fig. \ref{Fig2}). In simple words, $g(\mathcal{X}_i)$ can be thought as the global feature in the PIPN architecture shown in Fig. \ref{Fig2}. We further explain the role of shared MLPs and the symmetric function in the following. We define $n_s$ as a global scaling variable controlling the size of the PIPN. In the following, we denote batch size by $B$, and reemphasize that ``batch size'' in this study is the number of domains fed into PIPN at each epoch.

In practice, the input of PIPN is a three-dimensional tensor (i.e., numeric array) of size $B\times N \times 2$. Afterward, there are two sequential shared MLPs with sizes of $(n_s\times 64, n_s\times 64)$ and $(n_s\times 64, n_s\times 128, n_s \times 1024)$, as displayed in Fig. \ref{Fig2}. In the next step, the symmetric operator ($s$) forms the global feature with a size of $n_s \times1024$. As exhibited in Fig. \ref{Fig2}, the global feature is concatenated to the intermediate feature tensor of size $B\times N\times(n_s\times64)$, resulting in a tensor of size $B\times N \times(n_s \times 1088)$. Next, two other sequential shared MLPs, respectively, with sizes of $(n_s\times 512, n_s\times 256, n_s\times 128)$ and $(n_s\times 128, n_{\text{PDE}})$ operate in the PIPN architecture. The output of the previous action is a tensor of size $B\times N \times n_{\text{PDE}}$, as shown in Fig. \ref{Fig2}. Using automatic differentiation \cite{tensorflow2015-whitepaper}, the governing PDEs (Eqs. \ref{Eq1}--\ref{Eq2}) are built and fed into the PIPN loss function. Note that reasonable choices for $n_s$ could be 0.125, 0.25, 0.5, 1.0, 2.0, and 4.0 such that a positive integer number represents the size of resulting shared MLPs. \\

\noindent\textbf{2.2.1.1 Shared MLPs and symmetric functions}

\noindent The PIPN is invariant with respect to $N!$ permutations of the input vector (see Fig. \ref{Fig2}). In other words, if we randomly permute the input vector ($\mathcal{X}_i$), the constructed geometry of the domain does not change and thus the solution ($\mathcal{Y}_i$) should not change. PIPN becomes permutation invariant by means of two features: shared MLPs and symmetric functions.

The concept of shared MLPs should not be confused with regular fully connected layers or so-called ``dense'' layers in the TensorFlow \cite{tensorflow2015-whitepaper} terminology. Here we explain the concept of shared MLPs with a simple example. Let us consider the first MLP layer in the first branch of PIPN with a size of $(n_s \times 64, n_s \times 64)$ as shown in Fig. \ref{Fig2}. Let us take $n_s=1$ for a moment. More specifically, we focus on the first shared layer with a size of 64. The transpose of the input vector $\mathcal{X}_i$ can be written as

\begin{equation}
\label{Eq5}
    \mathcal{X}_i^{tr} = \begin{bmatrix}
     x_1 & x_2 & \dots & x_N \\
     y_1 & y_2 & \dots & y_N
    \end{bmatrix}.
\end{equation}
After applying the first shared layer to $\mathcal{X}_i$, the output is a matrix of size $64 \times N$ and can be written as

\begin{equation}
\label{Eq6}
    \begin{bmatrix}
\textbf{a}_{64 \times 1}^{(1)} & \textbf{a}_{64 \times 1}^{(2)} & \dots & 
\textbf{a}_{64 \times 1}^{(N)}
\end{bmatrix},
\end{equation}
where $\textbf{a}_{64 \times 1}^{(1)}$, $\textbf{a}_{64 \times 1}^{(2)}$, ..., $\textbf{a}_{64 \times 1}^{(N)}$ are vectors, which are computed as follows

\begin{equation}
\label{Eq7}
    \begin{matrix}
\textbf{a}_{64 \times 1}^{(1)} = \sigma \Bigl(\textbf{W}_{64 \times 2} \begin{bmatrix} x_1 \\ y_1 \end{bmatrix}+ \textbf{b}_{64 \times 1}\Bigl), \\
\textbf{a}_{64 \times 1}^{(2)} = \sigma \Bigl(\textbf{W}_{64 \times 2} \begin{bmatrix} x_2 \\ y_2 \end{bmatrix}+ \textbf{b}_{64 \times 1}\Bigl), \\
\vdots \\
\textbf{a}_{64 \times 1}^{(N)} = \sigma \Bigl(\textbf{W}_{64 \times 2} \begin{bmatrix} x_N \\ y_N \end{bmatrix}+ \textbf{b}_{64 \times 1}\Bigl),  \\
\end{matrix}
\end{equation}
where \textbf{W} and \textbf{b} are the shared weight matrix and bias vector, respectively. The nonlinear activation function is shown by $\sigma$, which acts elementwise. As can be realized from Eq. \ref{Eq7}, the same (shared) \textbf{W} and \textbf{b} are applied to each spatial point of the domain with the corresponding vector of $[x_j  \; \; y_j]^{tr}$, while $1\leq j \leq N$. That is why we call it shared MLPs. A similar procedure is conducted at the rest of the layers. By this strategy, it is immediately concluded that each point is independently processed in the PIPN framework and the only place that points meet each other is where the global feature needs to be identified (see Eq. \ref{Eq4}).

Concerning the symmetric function, we consider the two formats: one being the ``maximum'' function such that

\begin{equation}
\label{Eq8}
    g(\mathcal{X}_i)=\max(h(x_1,y_1 ),\dots,h(x_N,y_N)); \forall (x_j,y_j)\in \mathcal{X}_i \text{ with } 1 \leq i \leq m \text{ and } 1 \leq j \leq N,
\end{equation}
and the other being the ``average'' function such that

\begin{equation}
\label{Eq9}
    g(\mathcal{X}_i)=\text{average}(h(x_1,y_1 ),\dots,h(x_N,y_N)); \forall (x_j,y_j)\in \mathcal{X}_i \text{ with } 1 \leq i \leq m \text{ and } 1 \leq j \leq N.
\end{equation}
We compare the PIPN performance using these two symmetric functions. One may refer to Ref. \cite{kashefi2022physics} for a further description of the details of PIPN; and to Ref. \cite{qi2017pointnet} for a further discussion on the computer science aspects of PointNet \cite{qi2017pointnet}.

\subsubsection{Loss function}\label{Sect222}
For any pair of $\mathcal{X}_i$ and $\mathcal{Y}_i$ ($1 \leq i \leq m$), the residuals of the linear momentum in the $x$ direction ($\mathcal{J}_i^{{\text{momentum}_x}}$), linear momentum in the $y$ direction ($\mathcal{J}_i^{\text{momentum}_y}$), along with the residuals of the sparse observations of the displacement field ($\mathcal{J}_i^{\text{displacement}_{\text{sensor}}}$) are respectively defined as follows:

\begin{equation}
\label{Eq10}
\mathcal{J}_i^{{\text{momentum}_x}} = \frac{1}{N} \sum_{k=1}^{N}\biggl(-\frac{\delta}{\delta x_k}\Bigl( \frac{1}{1-\nu}\frac{\delta \Tilde{u}_k}{\delta x_k}+\frac{\nu}{1-\nu}\frac{\delta \Tilde{v}_k}{\delta y_k}\Bigl) -\frac{\delta}{\delta y_k}\Bigl(\frac{1}{2}\bigl(\frac{\delta \Tilde{u}_k}{\delta y_k} + \frac{\delta \Tilde{v}_k}{\delta x_k}\bigl)\Bigl) + \frac{\alpha}{1-\nu}\frac{\partial T_k}{\partial x_k}\biggl)^2,
\end{equation}

\begin{equation}
\label{Eq11}
\mathcal{J}_i^{{\text{momentum}_y}} = \frac{1}{N} \sum_{k=1}^{N}\biggl(-\frac{\delta}{\delta y_k}\Bigl( \frac{\nu}{1-\nu}\frac{\delta \Tilde{u}_k}{\delta x_k}+\frac{1}{1-\nu}\frac{\delta \Tilde{v}_k}{\delta y_k}\Bigl) -\frac{\delta}{\delta x_k}\Bigl(\frac{1}{2}\bigl(\frac{\delta \Tilde{u}_k}{\delta y_k} + \frac{\delta \Tilde{v}_k}{\delta x_k}\bigl)\Bigl) + \frac{\alpha}{1-\nu}\frac{\partial T_k}{\partial y_k}\biggl)^2,
\end{equation}

\begin{equation}
\label{Eq12}
\mathcal{J}_i^{\text{displacement}_{\text{sensor}}} = \frac{1}{M}\sum_{k=1}^M \biggl(\Bigl(\Tilde{u}_k - u_k^{\text{sensor}}\Bigl)^2+\Bigl(\Tilde{v}_k - v_k^{\text{sensor}}\Bigl)^2\biggl),
\end{equation}
where $M$ is the number of sensors located at each point cloud for a sparse measurement of the displacement fields. The automatic differentiation operator is shown by $\delta$. The $x$ and $y$ components of the displacement fields measured at the sensor locations are shown by $u_k^{\text{sensor}}$ and $v_k^{\text{sensor}}$, respectively, while the network outputs are denoted by $\Tilde{u}_k$ and $\Tilde{v}_k$. Note that because we assume that the temperature field is known to us, then the temperature gradient is also known. The hyperbolic activation function defined as 
\begin{equation}
\label{Eq13}
    \sigma(\gamma) = \frac{e^{2 \gamma} - 1}{e^{2 \gamma} + 1},
\end{equation}
is implemented in all the layers of PIPN, similar to Refs. \cite{kashefi2022physics,kashefi2022prediction}. Note that due to the presence of the second-order spatial derivative of the displacement fields in the governing PDEs (Eqs. \ref{Eq1}--\ref{Eq2}), the second-order derivative of the activation function used in PIPN must be well-defined.

As can be realized from Eq. \ref{Eq3}, because the displacement fields are a function of $g(\mathcal{X}_i)$, the spatial derivatives of the displacement also become a function of $g(\mathcal{X}_i)$ in PIPN. For example,

\begin{equation}
\label{Eq14}
    \frac{\delta u_j}{\delta x_j} = \frac{\delta f((x_j,y_j),g(\mathcal{X}_i))}{\delta x_j}; \forall (x_j,y_j) \in \mathcal{X}_i \text{ and } \forall u_j \in \mathcal{Y}_i \text{ with } 1 \leq i \leq m \text{ and } 1 \leq j \leq N.
\end{equation}
Similar expressions can be written for $\frac{\delta u_j}{\delta y_j}$, $\frac{\delta v_j}{\delta x_j}$, $\frac{\delta}{\delta x_j}(\frac{\delta u_j}{\delta x_j})$, etc. Hence, all the components presented in the loss function of PIPN contain the geometric information of the point clouds of the set $\Phi=\{V_i\}_{i=1}^m$, a specific feature that is not available in regular PINNs \cite{raissi2019physics}. Eventually, the PIPN loss function is written as

\begin{equation}
\label{Eq15}
    \mathcal{J} = \frac{B}{m} \sum_{b=1}^{m/B}\biggl(\frac{1}{B} \sum_{i=1+(b-1)B}^{bB} \Bigl(\omega_\text{momentum}\bigl(\mathcal{J}_i^{{\text{momentum}_x}} + \mathcal{J}_i^{{\text{momentum}_y}}\bigl) + \omega_\text{sensor}\mathcal{J}_i^{\text{displacement}_{\text{sensor}}}\Bigl)\biggl),
\end{equation}
where $\omega_\text{momentum}$ and $\omega_\text{sensor}$ are the corresponding weights of each residual, while their units are the inverse of the unit of their associated residuals such that the total loss function ($\mathcal{J}$) becomes unitless at the end. We elaborate on the choice of these weight factors in the following section. \\

\noindent\textbf{2.2.2.1 Component weights in the PIPN loss function}

\noindent Based on our machine learning experiments, the choice of $\omega_{\text{momentum}}$ and $\omega_{\text{sensor}}$ remarkably affect the convergence rate of the training procedure. Here, we propose a few simple, but practical functions for $\omega_{\text{momentum}}$ and $\omega_{\text{sensor}}$, and discuss their effects in Sect. \ref{Sect3}. The first trivial selection is setting an equal weight for both the momentum and sensor components of the loss function such that 

\begin{equation}
\label{Eq16}
    \begin{cases}
    \omega_\text{momentum} = 1 \text{ m} \\
    \omega_\text{sensor} = 1 \text{ m}^{-1}.
    \end{cases}
\end{equation}
The next choice is setting a higher weight for the sensor component compared to the momentum one in the PIPN loss function (see Eq. \ref{Eq13}) such that

\begin{equation}
\label{Eq17}
    \begin{cases}
    \omega_\text{momentum} = 1 \text{ m} \\
    \omega_\text{sensor} = \omega_0 \text{ m}^{-1} \text{, with } \omega_0 > 1.
    \end{cases}
\end{equation}
Note that $\omega_0$ remains constant during the training and can be thought of as a hyperparameter. Additionally, our machine learning experiments show that setting a higher weight for the momentum component compared to the sensor one leads to a significant reduction in the convergence speed and thus we do not propose it here. Our third proposal is setting a higher weight for the sensor component, while this weight exponentially decreases during the training process such that

\begin{equation}
\label{Eq18}
    \begin{cases}
    \omega_\text{momentum} = 1 \text{ m} \\
    \omega_\text{sensor} = \max \Bigl(\omega_1 \times \exp \bigl(\frac{\text{--epoch}}{r_1}\bigl), 1.0\Bigl) \text{ m}^{-1} \text{, with } \omega_1 > 1 \text{ and } r_1 > 0,  
    \end{cases}
\end{equation}
where ``epoch'' refers to the training iteration. Again, $\omega_1$ and $r_1$ are the hyperparameters. Our last suggestion is similar to our third one with the difference that the weight decreases logarithmically  such that 

\begin{equation}
\label{Eq19}
    \begin{cases}
    \omega_\text{momentum} = 1 \text{ m} \\
    \omega_\text{sensor} = \max \Bigl(\omega_2 \times \ln \bigl(\text{--epoch}+r_2\bigl), 1.0\Bigl) \text{ m}^{-1} \text{, with } \omega_2 > 1 \text{ and } r_2 > 0,  
    \end{cases}
\end{equation}
where $\omega_2$ and $r_2$ are the hyperparameters. Note that in all our proposals, $\omega_{\text{sensor}}$ never becomes less than one. Again, the units of $\omega_{\text{momentum}}$ and $\omega_{\text{sensor}}$ are set such that $\mathcal{J}$ becomes unitless.

At the end of this subsection, we address the fact that setting optimal weights in the loss functions of deep learning models is an active research area by itself (see e.g., Ref. \cite{xiang2021self}); however, this is the first time that we introduce this concept to the PIPN configuration.

\subsection{Data generation}\label{Sect23}

\begin{figure*}
\centering
\includegraphics[width=1.0\textwidth]{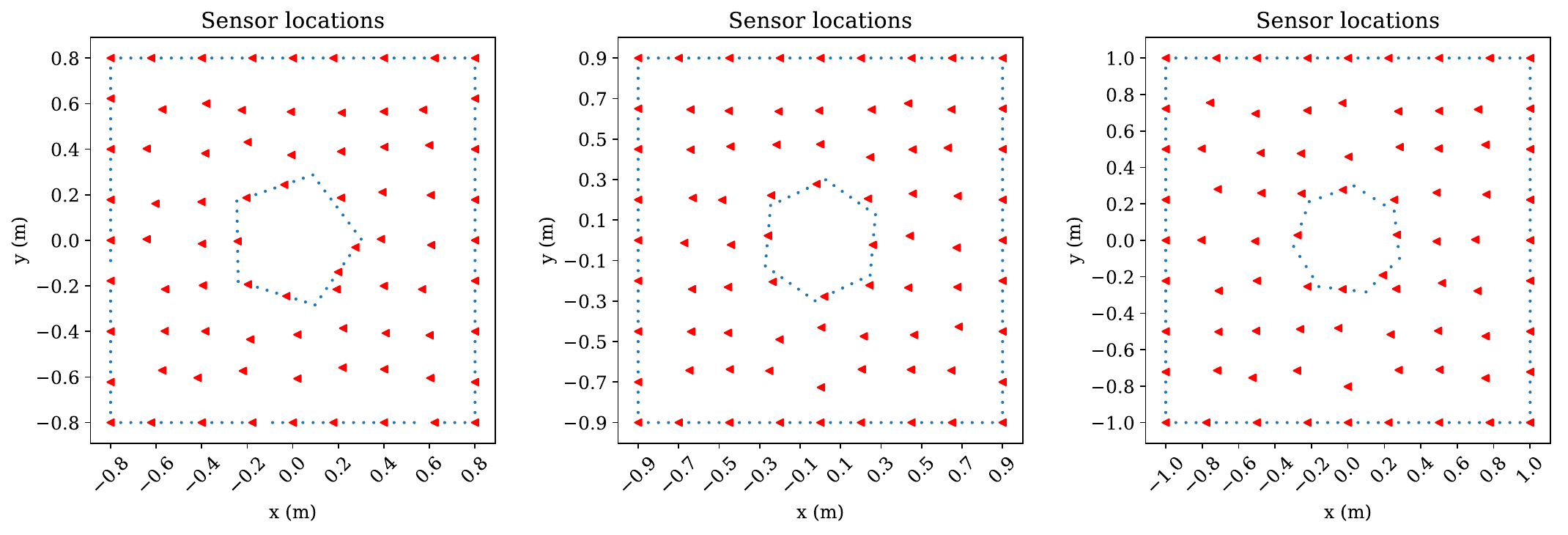}
\caption{Sensor locations for three domains from the set $\Phi=\{V_i\}_{i=1}^{532}$; Red triangles show sensors measuring the displacement fields}
\label{Fig3}
\end{figure*}

Computational domains of the set $\Phi=\{V_i\}_{i=1}^m$ are thin square plates with a cavity. The side length of the square plates takes three different values while the cavity takes shapes of a square, regular pentagon, regular hexagon, regular octagon, and regular nonagon. We further enlarge the number of domains by rotating the cavity with respect to the fixed thin square plates. Details of the geometric features of the set $\Phi=\{V_i\}_{i=1}^m$ are listed in Table \ref{Tab1}. In total, 532 geometries (i.e., $m=532$) are the input of PIPN. In this way, we establish the set $\Phi=\{V_i\}_{i=1}^{532}$. Three examples of geometries of this set are shown in Fig. \ref{Fig3}. Concerning the sensor locations, we initially assume the domains have no cavities and distribute sensors evenly within them. Any sensors located within the cavity are then relocated to the nearest point inside the domain. In our machine learning investigations, we consider batch sizes ($B$) of 7, 14, 19, 28, 38, 76, and 133, which are all divisors of 532. We recall that the maximum batch size ($B$) implemented by Kashefi and Mukerji \cite{kashefi2022physics} was 13 and the maximum number of input data ($m$) considered by Kashefi and Mukerji \cite{kashefi2022physics} was 108. Another difference is that the side length of the outer boundaries takes three values of 1.6 m, 1.8 m, and 2.0 m in the current study; however, Kashefi and Mukerji \cite{kashefi2022physics} exclusively considered a fixed value of 2.0 m for the side length of outer boundaries when they studied natural convection in a square enclosure with a cylinder (see Fig. 12 and Table 7 of Ref. \cite{kashefi2022physics}). In this way, we experience more variations in geometric features of the input point clouds in this study and thus a more challenging task is defined for PIPN.

The MATLAB PDE toolbox is used for the two purposes of validation of the PIPN predictions and generations of the sparse labeled data at virtual sensor locations. The ground truth data is generated using the toolbox as follows. The domains of the set $\Phi=\{V_i\}_{i=1}^{532}$ are loaded by thermal conduction with Dirichlet boundary conditions such that the temperature on the outer and inner boundaries are respectively set to 1 and zero in International Unit System. Note that the thermal conductivity does not affect the solution of the Laplace equation governing the steady-state thermal conduction. Moreover, we impose zero displacement boundary conditions on the boundaries of the domains of the set $\Phi=\{V_i\}_{i=1}^{532}$.

\begin{table}
\caption{Description of the generated geometries; $R$ indicates the radius of the circumscribed circle of the inner thin cavity. $\Omega$ denotes variation (in degree) in the orientation of the inner thin cavity with respect to its geometric center.}
\centering
\begin{tabular}{lllll}
\hline
\hline
\pbox{20cm}{The shape of the \\ inner thin cavity} & $R$ & $\Omega$  & \pbox{20cm}{Side length of thin \\ square plates} & \pbox{20cm}{Number of \\ selected data} \\
\hline
Square	& 0.35 m &	1, 3, \dots, 87, 89 	& 1.6 m, 1.8 m, 2 m &	134 \\
Regular pentagon & 0.30 m	& 1, 3, \dots, 69, 71 &	1.6 m, 1.8 m, 2 m &	107 \\
Regular hexagon	& 0.30 m &	1, 3, \dots, 57, 59 & 1.6 m, 1.8 m, 2 m & 90 \\
Regular heptagon & 0.30 m &	1, 3, \dots, 49, 51 & 1.6 m, 1.8 m, 2 m &	76 \\
Regular octagon	& 0.30 m & 1, 3, \dots, 43, 45 & 1.6 m, 1.8 m, 2 m	& 66 \\
Regular nonagon	& 0.30 m & 1, 3, \dots, 37, 39 & 1.6 m, 1.8 m, 2 m & 59 \\
\hline
Total & &  & & 532 \\
\hline
\hline
\end{tabular}
\label{Tab1}
\end{table}

%%%%%%%%%%%%%%%%%%

\subsection{Computational considerations}\label{Sect24}

The Poisson ratio ($\nu$) is set to 0.3 and the thermal expansion coefficient ($\alpha$) is set to 1.0 in the International Unit System. Note that elastic Young`s  modulus ($E$) does not appear in Eqs. (\ref{Eq10}--\ref{Eq11}) and thus is not a problem input. Furthermore, we set $N=2021$ and $M=81$ in the point clouds. The sensors are approximately equally spaced and located in the domains of the set $\Phi=\{V_i\}_{i=1}^{532}$. The sensor locations for three different domains taken from the set $\Phi=\{V_i\}_{i=1}^{532}$ are exhibited, for instance, in Fig. \ref{Fig3}. We use the Adam optimizer \cite{kingma2014adam} with the hyperparameters of $\beta_1=0.9$, $\beta_2=0.999$, and $\hat{\epsilon}=10^{-6}$. The mathematical explanation of $\beta_1$, $\beta_2$, and $\hat{\epsilon}$ are expressed in Ref. \cite{kingma2014adam}. The constant learning rate of 0.0003 is set for the entire machine learning experiments. For a fair comparison, all the simulations are run with double precision on an NVIDIA A100 SXM4 graphic card with a memory clock rate of 1.41 GHz and 40 Gigabytes of RAM.

\begin{figure*}[htp]
\centering
\includegraphics[width=1.0\textwidth]{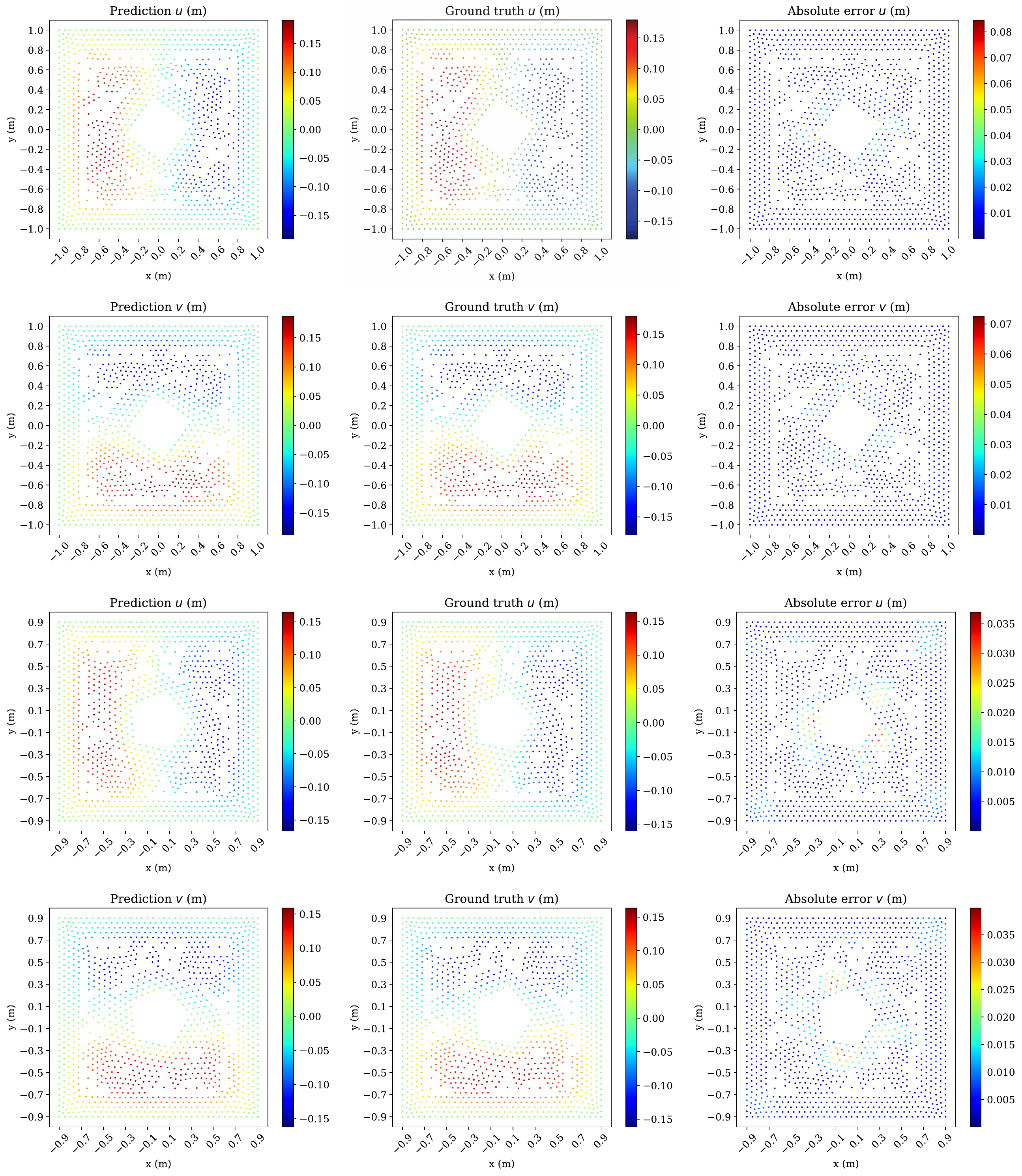}
\caption{The first group of examples taken from the set $\Phi=\{V_i\}_{i=1}^{532}$, comparing the finite element solutions to the PIPN predictions for the displacement fields}
\label{Fig4}
\end{figure*}

\section{Results and discussion}\label{Sect3}

\subsection{General analysis}\label{Sect31}

In Table \ref{Tab2}, we tabulate the average, maximum, and minimum errors of the predicted displacement fields in the $x$ and $y$ directions for the domains of the set $\Phi=\{V_i\}_{i=1}^{532}$ with batch size of $B=28$, the network size of $n_s=1.0$, and the weight of $\omega_{\text{sensor}}=50$ m$^{-1}$ for the sensor component in the PIPN loss function (see Eq. \ref{Eq15}). According to the data tabulated in Table \ref{Tab2}, the average pointwise relative errors ($L^2$ norm) for all 532 geometries of the set $\Phi$ is less than 9\%, indicating a successful accomplishment for the PIPN framework. Moreover, the maximum pointwise relative errors ($L^2$ norm) do not exceed 14\%, demonstrating a reasonable accuracy for engineering applications.

\begin{figure*}[htp]
\centering
\includegraphics[width=1.0\textwidth]{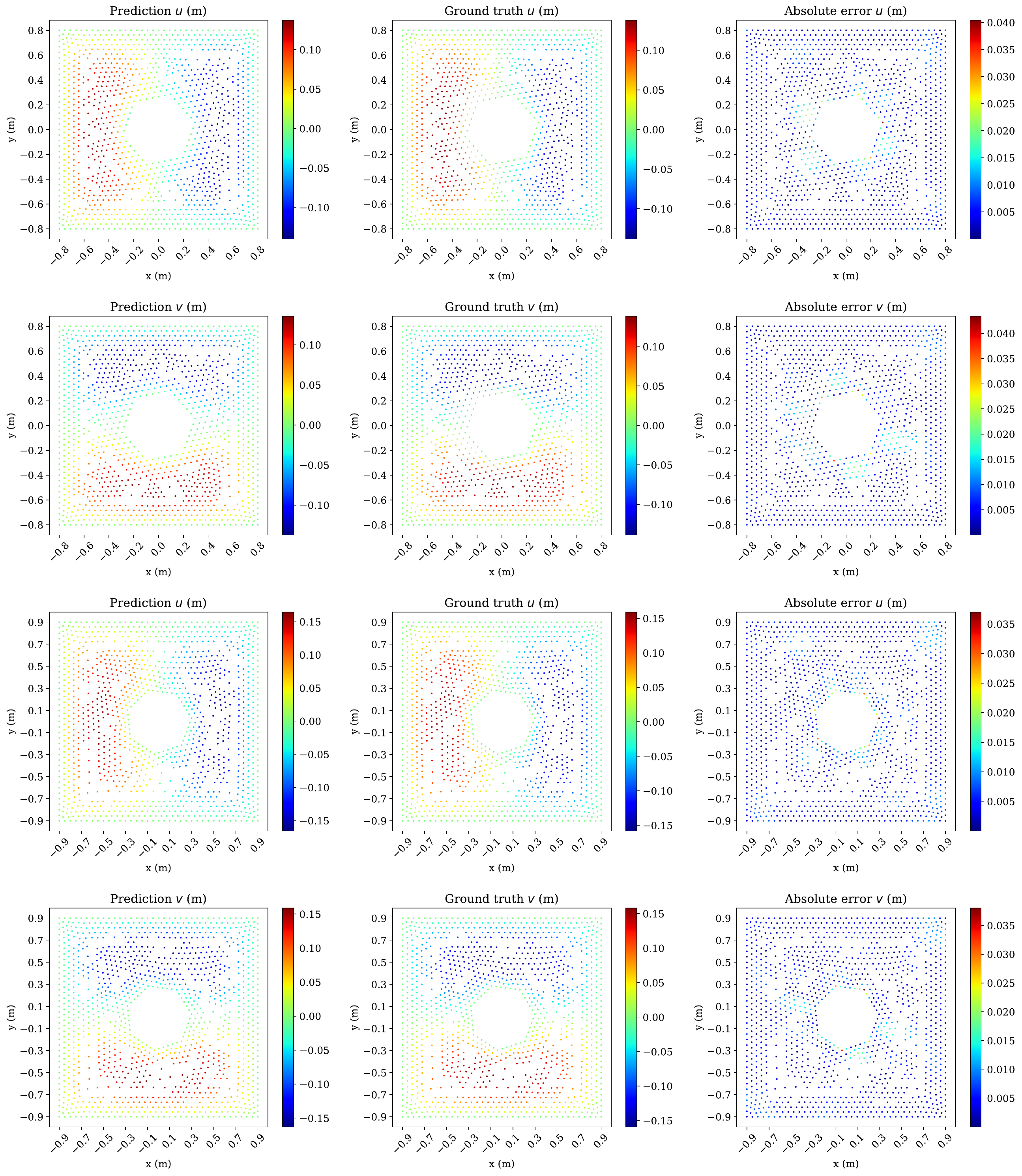}
\caption{The second group of examples taken from the set $\Phi=\{V_i\}_{i=1}^{532}$, comparing the finite element solutions to the PIPN predictions for the displacement fields}
\label{Fig5}
\end{figure*}

%%%%%%%

\begin{figure*}[htp]
\centering
\includegraphics[width=1.0\textwidth]{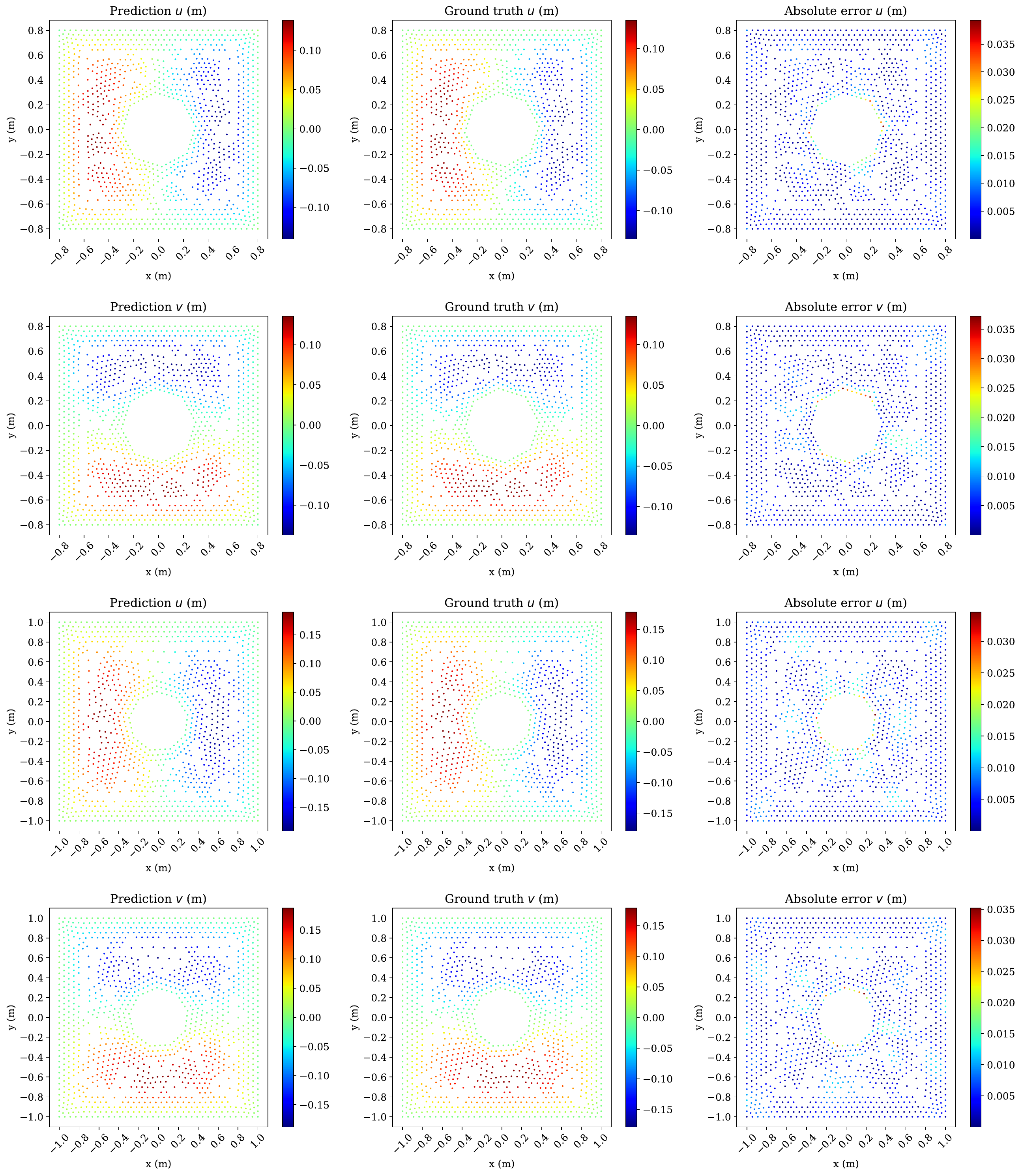}
\caption{The third group of examples taken from the set $\Phi=\{V_i\}_{i=1}^{532}$, comparing the finite element solutions to the PIPN predictions for the displacement fields}
\label{Fig6}
\end{figure*}

\begin{figure*}
\centering
\includegraphics[width=1.0\textwidth]{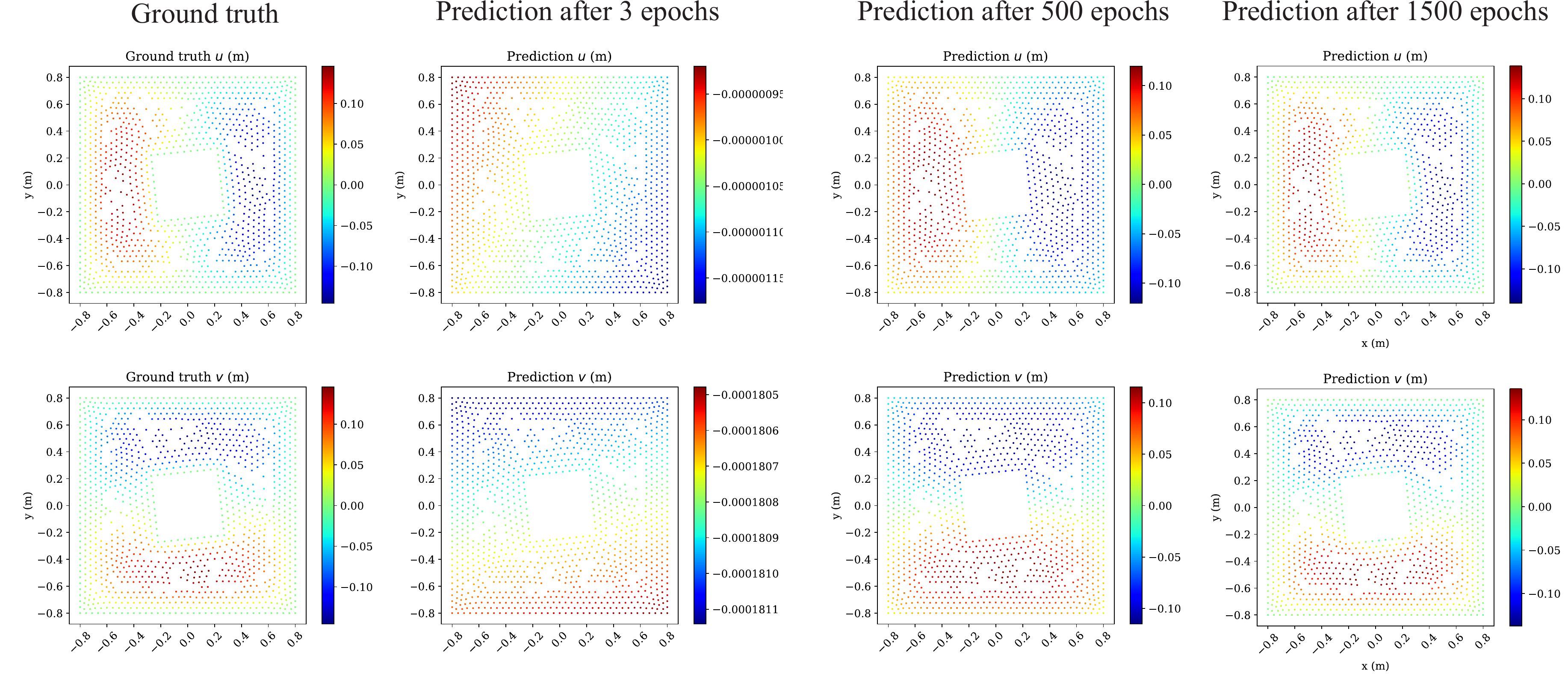}
\caption{A comparison between the ground truth and prediction of PIPN for the displacement fields after 3, 500, and 1500 epochs}
\label{Fig7}
\end{figure*}

A comparison between the ground truth (i.e., the finite element solutions) and the PIPN predictions is made in Figs. \ref{Fig4}--\ref{Fig6} for six different geometries taken from the set $\Phi=\{V_i\}_{i=1}^{532}$. As can be observed in Figs. \ref{Fig4}--\ref{Fig6}, there is an excellent agreement between the ground truth and the displacement fields predicted by PIPN. For each geometry, the maximum local pointwise error happens on the boundaries of the inner cavity of thin plates. This outcome is expectable because, first, PIPN is not informed from boundary conditions (i.e., zero displacements), and second, most variations from one geometry to another one take place on the boundaries of the domains of the set $\Phi=\{V_i\}_{i=1}^{532}$. In fact, predicting boundary values is the most difficult task for PIPN during the training procedure. This fact can be realized by looking at Fig. \ref{Fig7}. We exhibit the prediction of PIPN for the displacement fields for one geometry taken from the set $\Phi=\{V_i\}_{i=1}^{532}$ after 3, 500, and 1500 epochs. Based on what can be seen in Fig. \ref{Fig7}, the PIPN prediction is inaccurate after 10 epochs. After 500 epochs, the displacement fields are accurately predicted by PIPN except on the boundaries of the domain. An improvement in the prediction of zero displacements on the boundaries of the domain is observed at the PIPN outcome after 1500 epochs. Hence, we conclude that the most time-consuming part for PIPN is to predict the right values on the domain boundaries.

\begin{figure*}[t]
\centering
\includegraphics[width=0.8\textwidth]{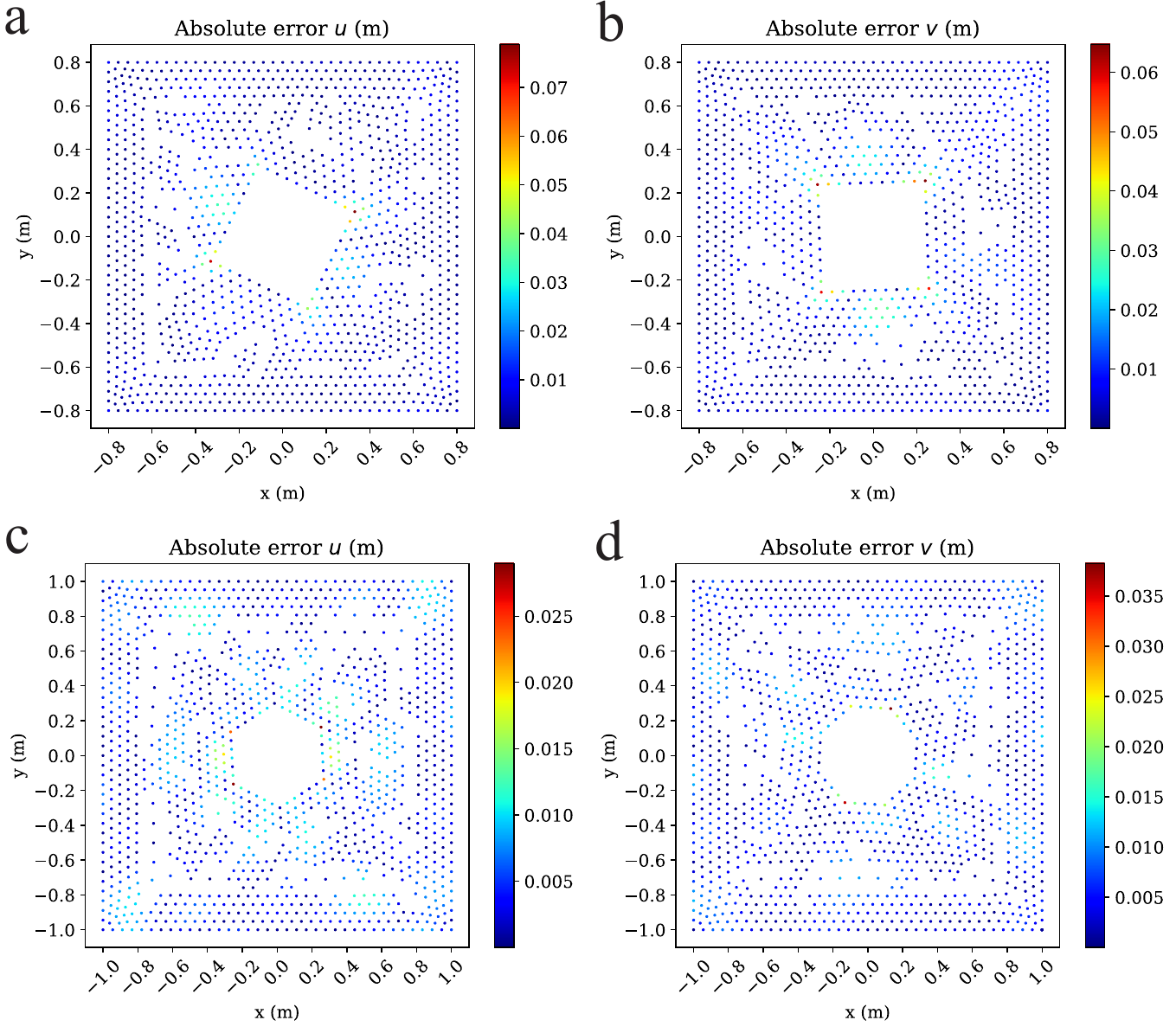}
\caption{Distribution of absolute pointwise error when the relative pointwise error ($L^2$ norm) becomes (\textbf{a}) maximum for $\tilde{u}$, (\textbf{b}) maximum for $\tilde{v}$, (\textbf{c}) minimum for $\tilde{u}$, and (\textbf{d}) minimum for $\tilde{v}$}
\label{Fig8}
\end{figure*}

The absolute pointwise error ($L^2$ norm) for geometries with the maximum and minimum relative pointwise errors of the displacement fields in the $x$ and $y$ directions are shown in Fig. \ref{Fig8}. According to Fig. \ref{Fig8}, the extremum errors occur for domains with different geometries of the set $\Phi=\{V_i\}_{i=1}^{532}$, revealing the fact that PIPN is not overfitted to one specific geometry. Moreover, we observe that the maximum pointwise errors happen for the domains with the side length of 1.6 m, which is the smallest side length available in the data set $\Phi=\{V_i\}_{i=1}^{532}$. On the other hand, the minimum pointwise errors occur for geometries with the side square length of 2.0 m, which is the largest available side length in the data set $\Phi=\{V_i\}_{i=1}^{532}$. Based on our deep learning experiments, the ideal range for the spatial coordinates of the input geometries is when they are expanded in [$-$1, 1]. Having stretched geometries out of [$-$1, 1] or compressed geometries inside of [$-$1, 1] results in an error increase. Therefore, the domains with a side length of 1.6 m end up having higher levels of errors compared to larger side lengths.

\begin{table}
\centering
\caption{Error analysis of the displacement fields for the domains of the set $\Phi=\{V_i\}_{i=1}^{532}$ when $B=28$, $n_s=1$, $\omega_{\text{sensor}}=50$ m$^{-1}$, and $\omega_{\text{momentum}}=1$ m. $||\dots||$ indicates the $L^2$ norm.}
\begin{tabular}{lll}
\hline
\hline
& $||\Tilde{u}-u||/||u||$ & $||\Tilde{v}-v||/||v||$ \\
\hline
Average & 8.11108E$-$2 & 8.60237E$-$2 \\
Minimum & 5.74907E$-$2 & 6.32261E$-$2 \\
Maximum & 1.19782E$-$1 & 1.31120E$-$1\\
\hline
\hline
\end{tabular}
\label{Tab2}
\end{table}

%%%%%%%%%%%

\begin{table}[H]
\centering
\caption{Error analysis of the displacement fields for the domains of the set $\Phi=\{V_i\}_{i=1}^{532}$ for different batch sizes ($B$) and three different network sizes of $n_s=1.0$, $n_s=0.5$, and $n_s=0.25$ when $\omega_{\text{sensor}}=50$ m$^{-1}$ and $\omega_{\text{momentum}}=1$ m. $||\dots||$ denotes the $L^2$ norm. The cross symbol ($\times$) indicates that the machine learning experiment is not doable due to a memory limitation.}
\begin{tabular}{lllllll}
\hline
\hline
Batch size $(B)$ &  $n_s=1.0$ &  $n_s=1.0$ & $n_s=0.5$ & $n_s=0.5$ & $n_s=0.25$ & $n_s=0.25$ \\
\hline
 & \pbox{20cm}{Average \\ $||\Tilde{u}-u||/||u||$}	& \pbox{20cm}{Average \\ $||\Tilde{v}-v||/||v||$} & \pbox{20cm}{Average \\ $||\Tilde{u}-u||/||u||$} &
 \pbox{20cm}{Average \\ $||\Tilde{v}-v||/||v||$} &
 \pbox{20cm}{Average \\ $||\Tilde{u}-u||/||u||$} &
 \pbox{20cm}{Average \\ $||\Tilde{v}-v||/||v||$} \\
\hline
7 &	8.56410E$-$2 & 8.58942E$-$2 & 8.40571E$-$2 & 8.90152E$-$2 & 1.00000 & 1.00000 \\
14 & 8.24455E$-$2 & 7.84251E$-$2 & 8.71806E$-$2 & 7.68771E$-$2 &	9.99999E$-$1 & 1.00000 \\ 
19 & 8.34842E$-$2 & 8.42466E$-$2 & 8.03585E$-$2 & 8.33552E$-$2 & 5.08574E$-$1 & 1.00000 \\
28 & 8.11108E$-$2 & 8.60237E$-$2 & 7.44200E$-$2 & 7.40376E$-$2 & 3.24522E$-$1 & 3.45531E$-$1  \\
38 & 8.69939E$-$2 & 9.05297E$-$2 & 8.45176E$-$2 & 8.20743E$-$2 & 3.68147E$-$1 & 3.69285E$-$1 \\
76	& $\times$ & $\times$ &	3.77084E$-$1 & 3.78790E$-$1 & 1.00000	& 9.99997E$-$1 \\
133	& $\times$ & $\times$ & $\times$ & $\times$ & 1.00000 & 5.05933E$-$1 \\
\hline
\hline
\end{tabular}
\label{Tab3}
\end{table}

%%%%%%%%%%%%%%%%

%\subsection{Effect of network size and batch size}\label{Sect32}

\subsection{Effect of batch size and network size}\label{Sect32}

We first explore the effect of batch sizes ($B$) on the PIPN performance in this subsection. It is worthwhile to note that at each epoch, the geometries are shuffled and randomly divided into mini-batch, exactly similar to the scenario of fully supervised learning models. We collect the average relative pointwise errors ($L^2$ norm) of the displacement fields for the batch sizes ($B$) of 7, 14, 19, 28, 38, 76, and 133 for three different network sizes ($n_s$) of 1.0, 0.5, and 0.25 in Table \ref{Tab3}. Note that due to the memory limitations, a few of these machine learning experiments are not doable such as the combination of a batch size of $B=133$ with the network size of $n_s=1.0$. According to Table \ref{Tab3}, for a fixed network size of $n_s=1.0$, the batch size ($B$) does not significantly affect the prediction accuracy. A similar story is true when the network size of $n_s=0.5$ is taken, except for the batch size of $B=76$, where the accuracy of the PIPN prediction notably decreases and the average relative pointwise errors ($L^2$ norm) of the displacement fields becomes approximately 37\%. The optimal batch size ($B$) is 28 for this network size ($n_s=0.5$). By reducing the network size to $n_s=0.25$, the overall performance of PIPN sharply decreases. For batch sizes ($B$) of 7, 14, 19, 76, and 133, the PIPN prediction is completely off. For the batch sizes ($B$) of 28 and 38, although the average relative pointwise errors ($L^2$ norm) are less than 50\%, the PIPN solution is not reliable. The lack of PIPN performance for the choice of $n_s=0.25$ is further discussed at the end of this subsection. But it is concluded that it is important to first select a PIPN with a suitable size, before investigating the effect of the batch size ($B$). In other words, the effect of network size on PIPN is significantly more noticeable compared to the influence of the batch size.

\begin{table}[H]
\centering
\caption{Error analysis of the displacement fields for the domains of the set $\Phi=\{V_i\}_{i=1}^{532}$ for different network sizes $(n_s)$ when $B=19$, $\omega_{\text{sensor}}=50$ m$^{-1}$, and $\omega_{\text{momentum}}=1$ m. $||\dots||$ indicates the $L^2$ norm.}
\begin{tabular}{lll}
\hline
\hline
Network size $(n_s)$ & Average $\frac{||\Tilde{u}-u||}{||u||}$ &  Average $\frac{||\Tilde{v}-v||}{||v||}$\\
\hline
0.125 & 1.00001	& 1.00000 \\
0.25 &  5.08574E$-$1 & 1.00000\\
0.5 &  8.03585E$-$2 & 8.33552E$-$2\\
1.0 &  8.34842E$-$2 & 8.42466E$-$2\\
1.5 &  8.51296E$-$2 & 7.35883E$-$2\\
2.0 &  9.25092E$-$2 &	1.00152E$-$1\\
2.5 &  1.51389E$-$1 &	1.52311E$-$1\\
\hline
\hline
\end{tabular}
\label{Tab4}
\end{table}

%%%%%%%%%%%%

\begin{figure*}[t]
\centering
\includegraphics[width=1.0\textwidth]{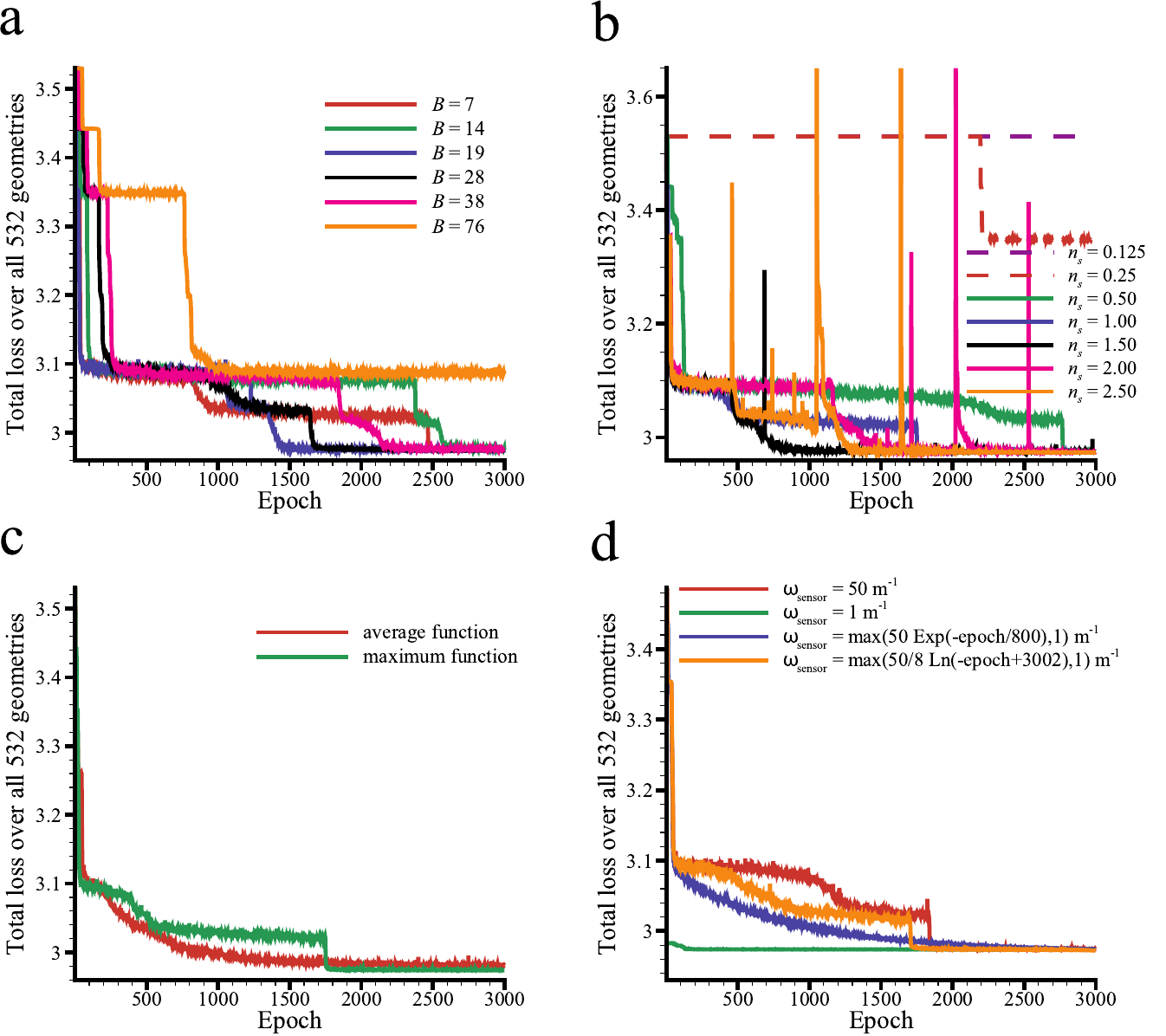}
\caption{Evolution of the PIPN loss function \textbf{a} for different batch sizes ($B$); \textbf{b} for different network sizes ($n_s$); \textbf{c} for two different symmetric functions (see Eqs. \ref{Eq8}--\ref{Eq9}); \textbf{d} for different setups of the weight of the sensor component in ($\omega_{\text{sensor}}$) in the loss PIPN function (see Eqs. \ref{Eq16}--\ref{Eq19})}
\label{Fig9}
\end{figure*}

The evolution of the PIPN loss function (see Eq. \ref{Eq15}) for the batch sizes ($B$) of 7, 14, 19, 28, 38, and 76 for the network size of $n_s=0.5$ is shown in Fig. \ref{Fig9}a. According to Fig. \ref{Fig9}a, for the batch sizes ($B$) of 7, 14, 19, 28, and 38, although the loss function eventually converges to approximately the same value, the machine learning experiment with the batch size of $B=19$ reaches this convergence with a significantly smaller number of epochs. Moreover, the PIPN loss associated with the batch size of $B=76$ converges to a value larger than the other investigated batch sizes, as can be realized from Fig. \ref{Fig9}a. This fact can be seen in another form by looking at the relative errors tabulated in Table \ref{Tab3} as already discussed.

Additionally, the effect of network size is carried out and outcomes are listed in Table \ref{Tab4}. Based on the information listed in Table \ref{Tab4}, for relatively small network sizes ($n_s=0.125$, $n_s=0.25$) and relatively large network sizes ($n_s=2.0$, $n_s=2.5$), the PIPN configuration experiences a higher level of errors compared to the moderate sizes of the network ($n_s=0.5$, $n_s=1.0$, $n_s=1.5$). For a relatively small PIPN, the network is too simple and a bias (i.e., underfitting) takes place. For a relatively large PIPN, the number of data is not sufficient to well determine the weight matrix (\textbf{W}) and bias vector (\textbf{b}) of the network, and hence, the PIPN predictions suffer from the lack of accuracy.

Furthermore, the evolution of the total loss value over all the geometries of the set $\Phi=\{V_i\}_{i=1}^{532}$ is displayed for the network sizes ($n_s$) of 0.125, 0.25, 0.5, 1.0, 1.5, 2.0, and 2.5 in Fig. \ref{Fig9}b. As can be seen in Fig. \ref{Fig9}b, for the network size ($n_s$) of 0.125, the training loss does not converge, leading to 100\% relative errors as tabulated in Table \ref{Tab4}. As discussed earlier, for the network sizes ($n_s$) of 0.5, 1.0, and 1.5, the relative errors of PIPN are approximately the same and smaller than other network size choices. Among these three selections, the network size ($n_s$) of 1.5 converges approximately after 800 epochs, while PIPN with the network size ($n_s$) of 0.5 converges after approximately 2800 epochs. For the network size ($n_s$) of 2.0 and 2.5, although the training loss converges, we observe a tremendous tendency for divergence and instability in the optimization process. The reason comes back to the fact that the number of data is not adequate for this choice of the PIPN size as explained in the previous paragraph.

\subsection{Choice of a symmetric function in PIPN}\label{Sect33}

We investigate the effectiveness of two different symmetric functions (see Eqs. \ref{Eq8}--\ref{Eq9}) in the setting of the PIPN configuration. Specifically, we set the batch size of $B=19$ and the network size of $n_s=1.0$ with the weight of $\omega_{\text{sensor}}=50$ m$^{-1}$ for the sensor component in the PIPN loss function for this machine learning experiment. Figure \ref{Fig9}c illustrates the evolution of the PIPN loss function (see Eq. \ref{Eq15}) for the maximum and average as two symmetric functions for extracting global features of point clouds. For a fair comparison, the network size and the batch size are set the same in both experiments. As can be observed in Fig. \ref{Fig9}c, the max function shows a better final performance compared to the average functions, though the loss with the average function initially drops faster than the loss function with the max function. However, a sharp decrease in the PIPN loss value with the maximum function occurs around approximately 1800 epochs such that this loss value becomes smaller than the corresponding value with the average function. Quantitatively, the average relative pointwise error ($L^2$ norm) of the displacement fields ($u$, $v$) in the $x$ and $y$ directions are respectively 4.69055E$-$1 and 4.39773E$-$1 for the average function while these errors are 8.34842E$-$2 and 8.42466E$-$2 for the maximum function.

%Figure \ref{Fig9}c exhibits the evolution of the PIPN loss for the different symmetric functions: ``average'' and ``maximum''. Accordingly, we observe convergence for both functions; however, the loss associated with the maximum function eventually converges to a lower value, leading to smaller relative errors as already reported in the previous paragraph. Interestingly, until approximately 1700 epochs, PIPN with the ``average'' function shows a better performance compared to the ``maximum'' function; however, a sharp decrease in the PIPN loss value with the ``maximum'' function occurs around approximately 1800 epochs such that this loss value becomes smaller than the corresponding value with the ``average'' function.

\begin{figure*}
\centering
\includegraphics[width=0.5\textwidth]{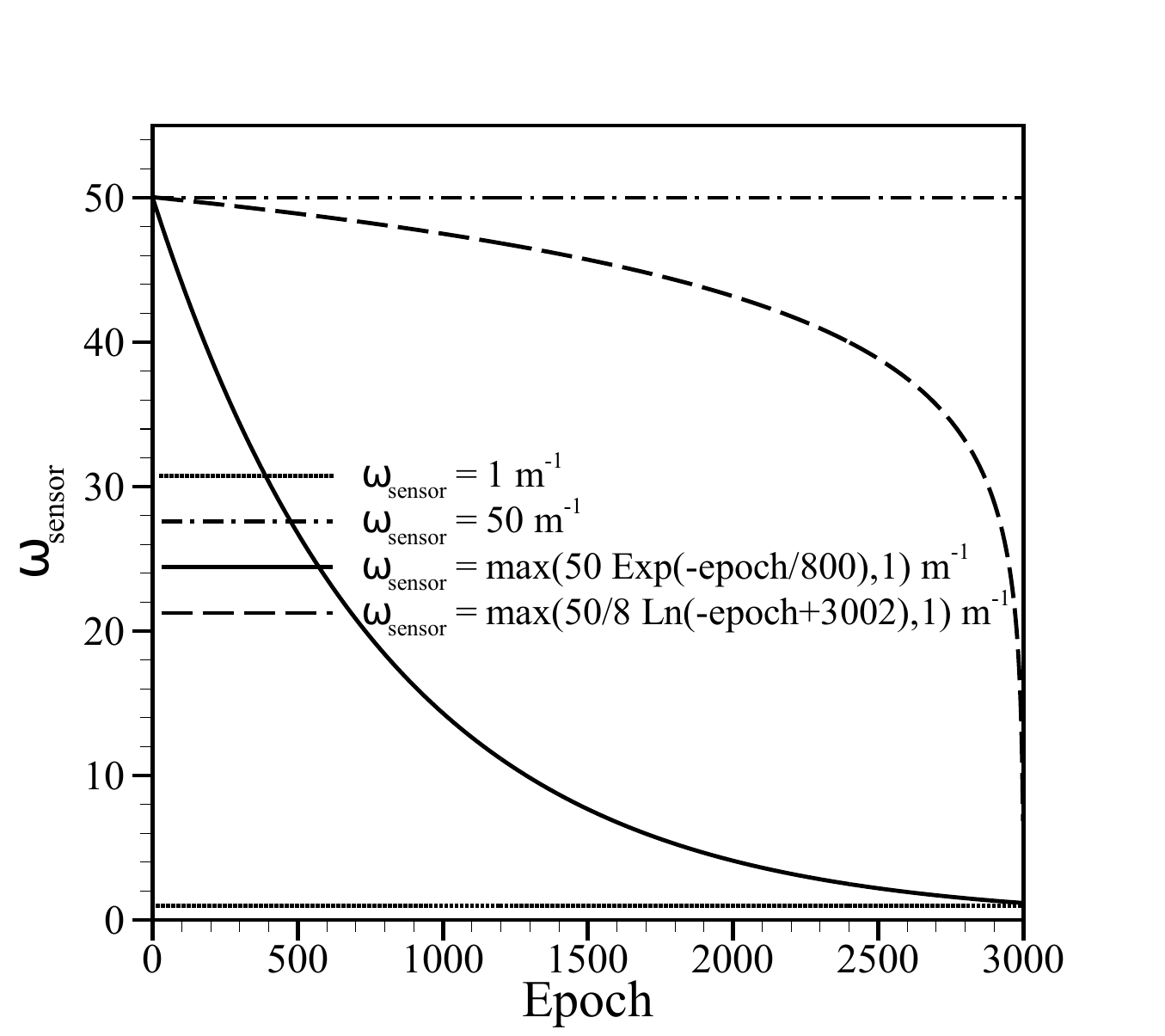}
\caption{Evolution of the weight of the sensor component ($\omega_{\text{sensor}}$) in the PIPN loss function during the training procedure for four different formulas (see Eqs. \ref{Eq16}--\ref{Eq19})}
\label{Fig10}
\end{figure*}

We note that Qi, et al. \cite{qi2017pointnet,qi2017pointnet++} used the maximum function for the purpose of object classification and segmentations in the area of computer graphics, and we observe here a higher performance with the maximum function in the area of computational mechanics as well. The reason can be explained as follows. As shown by Kashefi, et al. \cite{kashefi2021point} for supervised learning, using the maximum function, boundary points of point clouds of the domains always contribute to the global feature in PointNet \cite{qi2017pointnet}. This is while the most variation from one geometry to another one happens also on boundaries. In this way, PIPN can more clearly distinguish the PDE solution depending on the geometric features of each domain of the set $\Phi=\{V_i\}_{i=1}^{532}$.

\subsection{Effect of component weights in the PIPN loss function}\label{Sect34}

We investigate four different formulas for the weight component of the sensor data in the PIPN loss function. More specifically, we set $\omega_0=50$ m$^{-1}$ in Eq. \ref{Eq17}, we set $\omega_1=50$ m$^{-1}$ and $r_1=800$ in Eq. \ref{Eq18}, and we set $\omega_2=\frac{50}{8}$ m$^{-1}$ and $r_2=3002$ in Eq. \ref{Eq19}. We display the evolution of the weight of the sensor component ($\omega_{\text{sensor}}$) during training in Fig. \ref{Fig10} for these four different setups (Eqs. \ref{Eq16}--\ref{Eq19}). The relative pointwise errors ($L^2$ norm) are tabulated in Table \ref{Tab5}. In general, the best PIPN performance is when $\omega_{\text{sensor}}$ is larger than $\omega_{\text{momentum}}$ and decreases either logarithmically or not at all (see Eq. \ref{Eq17} and Fig. \ref{Fig9}d as well as Eq. \ref{Eq18} and Fig. \ref{Fig9}d).
%In general, the best PIPN performances are obtained when the sensor component ($\omega_{\text{sensor}}$) has a constant but higher value compared to the PDE component ($\omega_{\text{momentum}}$) (see Eq. \ref{Eq17} and Fig. \ref{Fig9}d) and when the sensor component ($\omega_{\text{sensor}}$) logarithmically decreases while still has a significantly higher value compared to the PDE component ($\omega_{\text{momentum}}$) during most of the training procedure (see Eq. \ref{Eq18} and Fig. \ref{Fig9}d).

Additionally, we plot the evolution of the PIPN loss function during training as a result of these four different formulas (see Eqs. \ref{Eq16}--\ref{Eq19}). These results demonstrate that setting a reasonably higher weight for the sensor component of the loss function (i.e., $\omega_{\text{sensor}}$) compared to the PDE component of the loss function (i.e., $\omega_{\text{momentum}}$) even during the entire training procedure leads to more accurate outputs. Mathematically, because PIPN is not informed of any boundary conditions in this problem, there are infinite solutions to the governing PDEs and thus obtaining the right solution is challenging for PIPN. But by setting a higher weight for the sensor data, PIPN is forced to have a priority for satisfying the mismatch between the network output and sensor data; however, this enforcement makes it convenient to also satisfy the governing PDEs because PIPN looks for the PDE solutions in the space created by the sensor data (rather than a weakly unknown space).

\begin{table}[H]
\centering
\caption{Error analysis of the displacement fields for the domains of the set $\Phi=\{V_i\}_{i=1}^{532}$ for different settings of weights $\omega_{\text{sensor}}$ in the loss PIPN function when $B=28$ and $\omega_{\text{momentum}}=1$ m. $||\dots||$ shows the $L^2$ norm.}
\begin{tabular}{cll}
\hline
\hline
$\omega_{\text{sensor}}$ & Average $\frac{||\Tilde{u}-u||}{||u||}$ &  Average $\frac{||\Tilde{v}-v||}{||v||}$\\
\hline
1.0 m$^{-1}$ & 3.72051E$-$1 & 3.75747E$-$1 \\
50.0 m$^{-1}$ & 8.11108E$-$2 & 8.60237E$-$2 \\
 $ \max \Bigl(50 \times \exp \bigl(\frac{\text{--epoch}}{800}\bigl), 1.0\Bigl)$ m$^{-1}$ & 3.71722E$-$1 & 3.71195E$-$1 \\
$\max \Bigl(\frac{50}{8} \times \ln \bigl(\text{--epoch}+3002\bigl), 1.0\Bigl)$  m$^{-1}$ & 9.81245E$-$2 & 9.63949E$-$2 \\
\hline
\hline
\end{tabular}
\label{Tab5}
\end{table}

Additionally, we plot the evolution of the total PIPN loss over all geometries of the set $\Phi=\{V_i\}_{i=1}^{532}$ for four different setups of the sensor component weight (see Eqs. \ref{Eq16}--\ref{Eq19}) in Fig. \ref{Fig9}d. Comparing the deep learning experiment with the constant value of $\omega_{\text{sensor}}=50$ m$^{-1}$ (see Eq. \ref{Eq17}) and the logarithmic function (see Eq. \ref{Eq19}), a convergence with fewer epochs happens for the logarithmic function, as can be seen in Fig. \ref{Fig9}d. Note that according to Table \ref{Tab5}, PIPN demonstrates a higher performance for the two mentioned choices of $\omega_{\text{sensor}}$ in comparison with the constant value of $\omega_{\text{sensor}}=1$ m$^{-1}$ (see Eq. \ref{Eq16}) and the exponential function (see Eq. \ref{Eq18}). Moreover, the PIPN loss evolution with the exponential function for presenting $\omega_{\text{sensor}}$ (see Eq. \ref{Eq18}) has a smooth decay and converges to a value, which is slightly higher than the convergence values of deep learning experiments with the constant value of $\omega_{\text{sensor}}=50$ m$^{-1}$ (see Eq. \ref{Eq17}) and the logarithmic function (see Eq. \ref{Eq19}). For the choice of the constant value of $\omega_{\text{sensor}}=1$ m$^{-1}$ (see Eq. \ref{Eq16}), the initial loss is smaller than the other three options simply because the weight of the sensor component in the loss function (see Eq. \ref{Eq15}) is scaled down by a factor of 50. However, the initial loss decreases with a very gentle slope up to 3000 epochs and thus there is a 38\% average relative error as reported in Table \ref{Tab5}.

\section{Summary and conclusions}\label{Sect4}

Generally speaking, there are two different deep learning categories in the area of computational mechanics: weakly supervised learning models, requiring sparse labeled data (see e.g., \cite{raissi2019physics}) and fully supervised learning models, requiring plentiful labeled data \cite{thuerey2019deep}. A specific class of supervised learning models is physics-informed neural networks (PINNs) \cite{raissi2019physics}. From an industrial point of view, an ideal deep learning framework should contain the ability to predict desired fields over hundreds, or even thousands, of domains with various geometries for the goal of swiftly optimizing geometric designs. Regular PINNs are applicable to the prediction of desired fields on a single geometry, whereas fully supervised learning models are used to predict desired fields on a few hundred geometries. In this sense, there is a gap between the weakly supervised learning and supervised learning models. Physics-informed PointNet (PIPN) \cite{kashefi2022physics} is a novel class of physics-informed deep learning algorithms that fills this gap. PIPN requires sparse label data but potentially is able to predict desired fields on a few hundred geometries.

In 2022, Kashefi and Mukerji \cite{kashefi2022physics} proposed PIPN and employed it to solve incompressible flow and thermal fields over 108 domains with different geometries, simultaneously. Furthermore, they only investigated the batch sizes of 1, 2, 3, and 4 for the natural convection problem and the batch size of 13 for the method of manufactured solutions \cite{kashefi2022physics}. In the current article, we tried for the first time to explore the underlying capacity of PIPN in terms of the number of geometries that desired fields can be predicted on, simultaneously, and also to investigate if PIPN (as a weakly supervised learning model) is able to compete with fully supervised learning models from this point of view or not. We answered this question by considering a linear elasticity problem and more specifically plane stress conditions.

Given our computational resources, we showed that PIPN was able to successfully predict the displacement fields over 532 domains with different geometries, simultaneously. The average relative pointwise error ($L^2$ norm) was approximately 9\% over the set. For the first time, we comprehensively explored the effects of batch size on the PIPN performance. By the term “batch size”, we meant the number of geometries that were fed into PIPN at each sub-epoch. Particularly, we executed batch sizes of 7, 14, 19, 28, 38, 76, and 133. In addition, we pioneered introducing a global parameter for controlling the network size of PIPN. Moreover, we systematically investigated for the first time the effect of this parameter on the accuracy of the PIPN predictions. It was concluded that the network size plays a more important role in controlling the PIPN performance compared to the batch size. It was observed that the network size should be compatible with the size of data (i.e., the number of geometries). In fact, we realized that when a suitable size for PIPN was selected, the effect of batch size on the output predictions by PIPN was insignificant. However, the batch size affected the convergence rate. Furthermore, we studied for the first time the accuracy of the displacement fields predicted by the PIPN methodology as a result of different constant weights and dynamics weights (i.e., as a function of epoch) for the component of the partial differential equations and the component of the sparse labeled data in the PIPN loss function. It was concluded that setting a constant higher weight for the component of the sparse labeled data compared to the component of the partial differential equations leads to higher prediction accuracy.

%\noindent
%\verb|\citet| $\rightarrow$ \citet{jenset&mcgil}\\
%\verb|\citet| $\rightarrow$ \citet{australiashealth}\\
%\verb|\citet| $\rightarrow$ \citet{shree-a}\\
%\verb|\citep| $\rightarrow$ \citep{fabricius-hansen2012b}\\
%\verb|\citealp| $\rightarrow$ (\citealp{eckhoff2018a})\\
%\verb|\citealp| $\rightarrow$ (\citealp{eckhoff2018a}; \citealp{fabricius-hansen2012b}; \citealp{shree-a})\\

%%%%%%%%%%%

\section*{Declaration of competing interest} 
The authors declare that they have no known competing financial interests or personal relationships that could have appeared to influence the work reported in this paper.

\section*{Data availability}
The data and the developed Python code are available on the following GitHub repository: 
\url{https://github.com/Ali-Stanford/PhysicsInformedPointNetElasticity}. One may also generate the code using ChatGPT \cite{kashefi2023chatgpt} by providing a description of the PIPN architecture detailed in this article.

\section*{Acknowledgements}
The authors acknowledge funding by the Shell-Stanford Collaborative Project on Digital Rock Physics 2.0 for supporting this research project. Additionally, we would like to thank the Stanford Research Computing Center for supporting our studies by providing computing resources. We wish to also thank the reviewers for their insightful comments and suggestions.

\bibliographystyle{plain}
\bibliography{bib}

\end{document}